\documentclass{article}

 \usepackage{graphicx}

 \usepackage[main, final]{neurips_2025}

\usepackage{listings}
\usepackage[utf8]{inputenc} 
\usepackage[T1]{fontenc}    
\usepackage{hyperref}       
\usepackage{url}            
\usepackage{booktabs}       
\usepackage{amsfonts}       
\usepackage{nicefrac}       
\usepackage{microtype}      
\usepackage{xcolor}         
\usepackage{amsmath}        
\usepackage{algorithm}      
\usepackage{algorithmic}    
\usepackage{subcaption}
\usepackage{gensymb}

\title{Assessing Situational and Spatial Awareness of VLMs with Synthetically Generated Video}

\author{%
  Pascal Benschop \\
  Department of Computer Science \\
  Delft University of Technology \\
  Delft \\
  \texttt{P.Benschop@tudelft.nl} \\
  \And
  Justin Dauwels \\
  Department of Computer Science \\
  Delft University of Technology \\
  Delft \\
  \texttt{J.H.G.Dauwels@tudelft.nl}
  \And
  Jan van Gemert\\
  Department of Computer Science \\
  Delft University of Technology \\
  Delft \\
  \texttt{J.C.vanGemert@tudelft.nl}
}

\begin{document}

\maketitle


\section*{Abstract}
Spatial reasoning in vision--language models (VLMs) remains fragile when semantics hinge on subtle temporal or geometric cues. We introduce a synthetic benchmark that probes two complementary skills: \emph{situational awareness} (recognizing whether an interaction is harmful or benign) and \emph{spatial awareness} (tracking who does what to whom, and reasoning about relative positions and motion). Through minimal video pairs, we test three challenges: distinguishing violence from benign activity, binding assailant roles across viewpoints, and judging fine-grained trajectory alignment. While we evaluate recent VLMs in a training-free setting, the benchmark is applicable to any video classification model. Results show performance only slightly above chance across tasks. A simple aid---stable color cues---partly reduces assailant role confusions but does not resolve the underlying weakness. By releasing data and code, we aim to provide reproducible diagnostics and seed exploration of lightweight spatial priors to complement large-scale pretraining.\footnote{\href{https://huggingface.co/datasets/PBenschop/Spatial_Reasoning_with_Synthetic_Video}{Hugging Face dataset} \;|\; \href{https://github.com/pascalbenschopTU/VLLM_AnomalyRecognition}{Code repository}.}

\section{Introduction}
Vision–language models can caption, answer questions, and summarize short videos, but \emph{spatial} and \emph{situational} awareness remain weak. By spatial awareness, we mean locating entities over time, understanding who does what to whom, and inferring relative directions and positions. By situational awareness, we mean recognizing the nature of an interaction—such as harmful versus benign—when appearance and context stay the same. These skills are critical for embodied assistance, mixed reality, collaborative robotics, and autonomy, where the question is not just what happens, but who acts, how, and with respect to whom.

Controlled, minimal-change stimuli can expose specific weaknesses that large, varied datasets might hide~\cite{johnson2017clevr, thrush2022winoground}. However, most prior work remains image-based or focuses on object permanence and occlusion in synthetic videos such as \textsc{CATER}~\cite{girdhar2020cater}. These do not target viewpoint-sensitive role binding or temporal geometry.

To address this, we build a compact synthetic benchmark that changes only a single spatial variable at a time while keeping appearance and background constant. This enables precise minimal-pair testing, discourages reliance on language shortcuts, and emphasizes consistency across pairs. Our goal is to pinpoint perceptual failures, distinguish them from language confusion, and inspire researchers to design structural biases that improve generalization.

We ask:
\begin{itemize}
  \item How consistently can VLMs show situational and spatial awareness across different scenes?
  \item How does the camera viewpoint affect tasks like role binding and motion alignment?
  \item Does prompting models to explain intermediate reasoning help their performance?
\end{itemize}

Our contributions include: realistic, synthetic video data with controllable variables; a minimal-pair benchmark targeting three core spatial tasks; and a training-free evaluation method mapping model outputs to labels via a simple text classifier.

\section{Related Work}

Synthetic–image benchmarks such as \textsc{CLEVR} pioneered tightly controlled scene graphs and program-grounded queries that expose compositional reasoning gaps \cite{johnson2017clevr}. Follow-up suites—including \emph{Is a Picture Worth a Thousand Words?}, \emph{Mind the Gap}, and \textsc{SpatialViz-Bench}—scale this paradigm with automatic template generation to probe a wider set of spatial relations, still restricted to static images \cite{Jiayu2024Picture,stogiannidis2025mind,wang2025spatialvizbenchautomaticallygeneratedspatial}.

Video-based benchmarks extend synthetic control into the temporal domain. \textsc{CATER} introduces occlusions, compositional actions, and viewpoint changes to test object permanence and spatio-temporal reasoning \cite{girdhar2020cater}.

Natural-image datasets aim for higher ecological validity while preserving structural supervision or minimal-pair design. \textsc{NLVR2} (paired photographs), \textsc{GQA} (scene-graph queries), and \textsc{Winoground} (hard caption–image pairs) highlight biases, consistency gaps, and visio–linguistic binding failures in realistic imagery \cite{suhr2019nlvr2,hudson2019gqa,thrush2022winoground}.

In parallel, VLMs have advanced rapidly by combining temporal video encoders with instruction-tuned LLMs. Representative families include Video-LLaMA and successors with improved spatio–temporal connectors, LLaVA-Video/Video-LLaVA with large-scale video instruction tuning, and Video-ChatGPT variants enabling open-ended video dialogue \cite{videollama2,VideoLLaMA3,gemma3,nvila,Qwen2.5-VL}. While these systems demonstrate strong generative and dialogue capabilities, systematic evaluations of viewpoint-relative reference, role binding, and heading geometry remain limited.

Our benchmark differs by retaining human actors and realistic surveillance backdrops while preserving the strict control of minimal-pair design \emph{across time}. Unlike static-image evaluations, each query spans multiple frames, testing whether models consistently track roles (e.g., \textit{assailant} vs.\ \textit{victim}) and spatial relations throughout an unfolding event—an ability that remains underexplored in existing benchmarks and VLM evaluations.

\section{Method}
\label{sec:method}
We generate short clips of 120 frames (30\,FPS) in Unreal Engine~5.5. Street scenes come from \href{https://dev.epicgames.com/documentation/en-us/unreal-engine/city-sample-project-unreal-engine-demonstration}{CitySample}; we additionally render a second background using an outdoor HDRI to vary global appearance while preserving geometry. Two crowd characters are positioned so that they initially face each other. Animations for fighting, dancing, idle, and walking are gathered from \href{https://www.mixamo.com/}{Mixamo} (see Appendix~\ref{app:unreal_setup}) and retargeted to the skeleton used by the CitySample characters. 

\begin{figure}[h]
    \centering
    \includegraphics[width=0.75\linewidth]{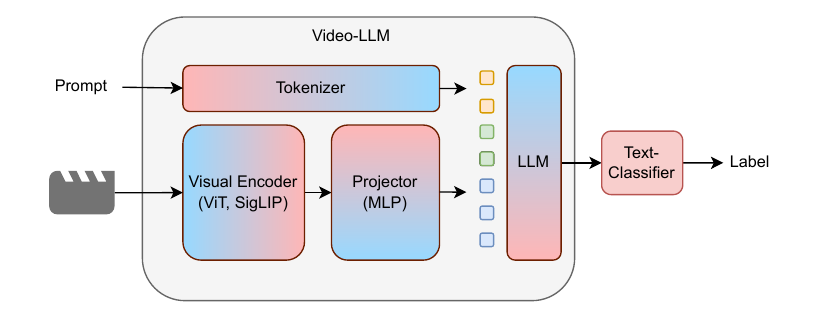}
    \caption{Evaluation pipeline. Image models receive uniformly sampled frames; video models receive the full clip. Free-form generations are mapped to the label set by a lightweight text classifier.}
    \label{fig:vlm_pipeline}
\end{figure}

We conduct experiments across two distinct backgrounds—a CitySample street scene and an outdoor HDRI-based environment—to improve the generalizability of our findings across different visual contexts. Within the CitySample scene, we introduce four camera variations: three fixed viewpoints (left, middle, right sidewalk positions) and one zoomed perspective (via focal-length adjustment). These variations add controlled visual diversity without altering scene geometry. 

Each task is cast as a binary classification from video to a small label set. Because free-form generations may not begin with the label token, we do not require single-token outputs. Instead, we map the model’s generation to the label set with a lightweight text classifier that receives the VLM response and the label schema (Figure~\ref{fig:vlm_pipeline}). Concretely, we use BART~\cite{BART} (\href{https://huggingface.co/facebook/bart-large-mnli}{facebook/bart-large-mnli}) to score candidate labels and select the best one. This rendering-and-evaluation setup lets us manipulate a single spatial factor while holding all others fixed.

\section{Experiments and Results}
We evaluate three core manipulations, each constructed as matched minimal pairs so that only one spatial factor changes:
\emph{(i) Situational discrimination} by varying motion style,
\emph{(ii) Role binding} by switching the attacker side,
and \emph{(iii) Following} by changing the follower’s heading offset to \( \pm 30^\circ \) relative to the leader.%
\footnote{We highlight the explicit heading offset of \( \pm 30^\circ \) to separate directional alignment from mere co-motion magnitude.}
In addition, we include a diagnostic \emph{color-identifier} variant for role binding, where stable clothing colors (left = red, right = blue) provide persistent role markers to test whether structural cues improve binding (see Appendix~\ref{app:color}, Figure~\ref{fig:colored_spatial_example}).

\begin{figure}[ht]
    \centering
    \includegraphics[width=1.0\linewidth]{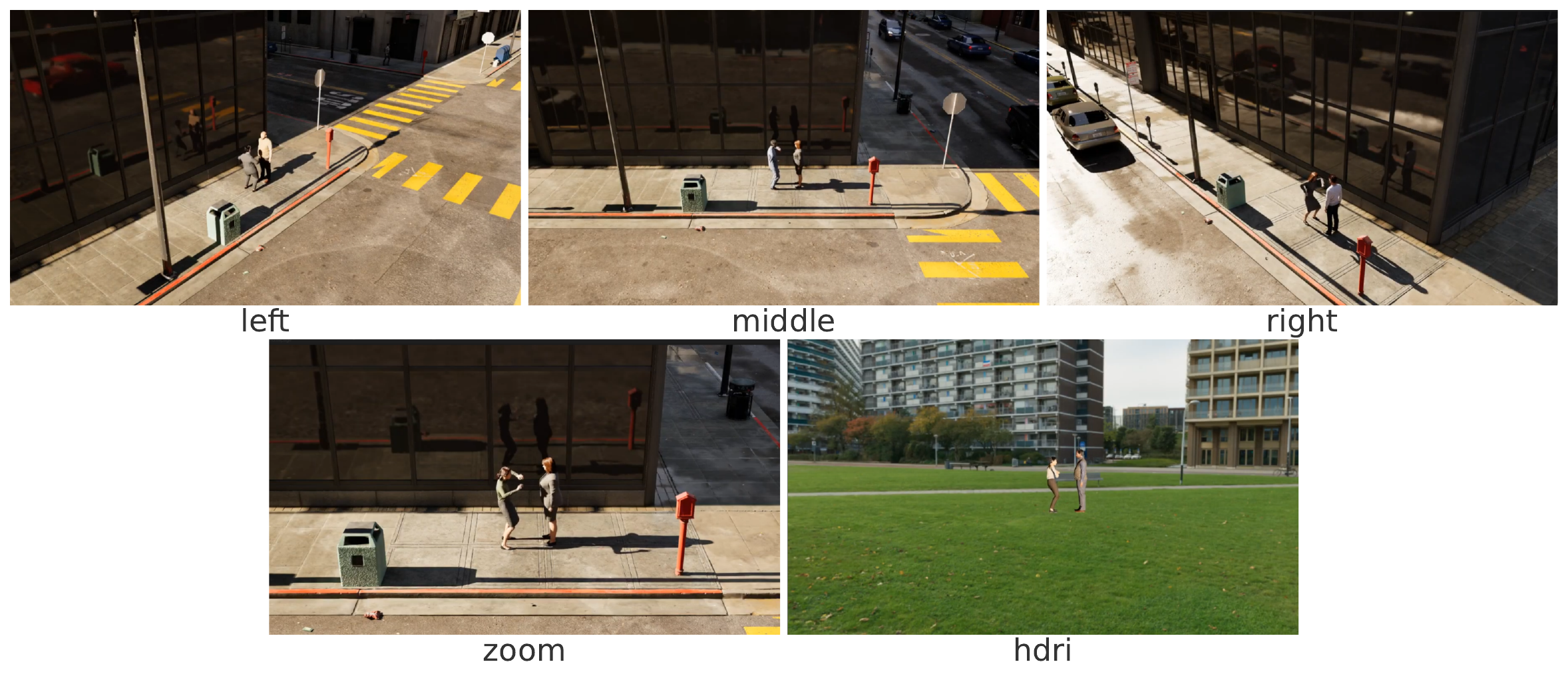}
    \caption{Example scenes used in experiments. The rightmost column uses the HDRI \emph{Buikslotermeerplein} for global illumination variation while preserving geometry.}
    \label{fig:example_scenes}
\end{figure}

Per experiment per class, we generate 40 videos: 8 variants per scene (with randomized characters each time) across 5 distinct scenes (see Figure~\ref{fig:example_scenes}). All tasks are run under the rendering and camera settings defined in Section~\ref{sec:method}. We evaluate several recent VLMs in a training-free setup with fixed prompts (see Appendix~\ref{app:prompts}).

Accuracy is reported as the fraction of correctly classified videos. Since classes are balanced, overall and macro accuracy coincide:
\[
\text{Acc} = \frac{1}{N} \sum_{i=1}^N \mathbf{1}\{\hat{y}_i = y_i\},
\]
where $N$ is the total number of videos, $y_i$ the ground-truth label, and $\hat{y}_i$ the predicted label.

\begin{figure}[h]
    \centering
    \includegraphics[width=0.99\linewidth]{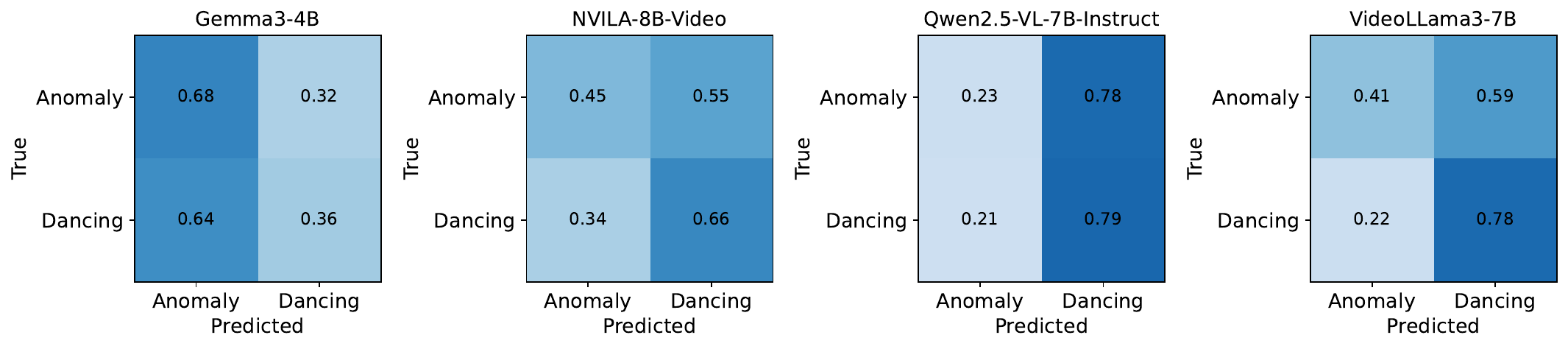}
    \caption{Situational discrimination (fighting vs.\ dancing/idle). Confusion matrices by model indicate reliable separation under controlled appearance and backgrounds.}
    \label{fig:anomaly_confusion}
\end{figure}

Figure~\ref{fig:anomaly_confusion} reports confusion matrices for situational discrimination. Only NVILA-8B-Video and VideoLLaMA3-7B reliably separate anomalies from dancing; Gemma3-4B over-flags anomalies and Qwen2.5-VL-7B over-predicts benign motion. 

\begin{figure}[h]
    \centering
    \includegraphics[width=0.9\linewidth]{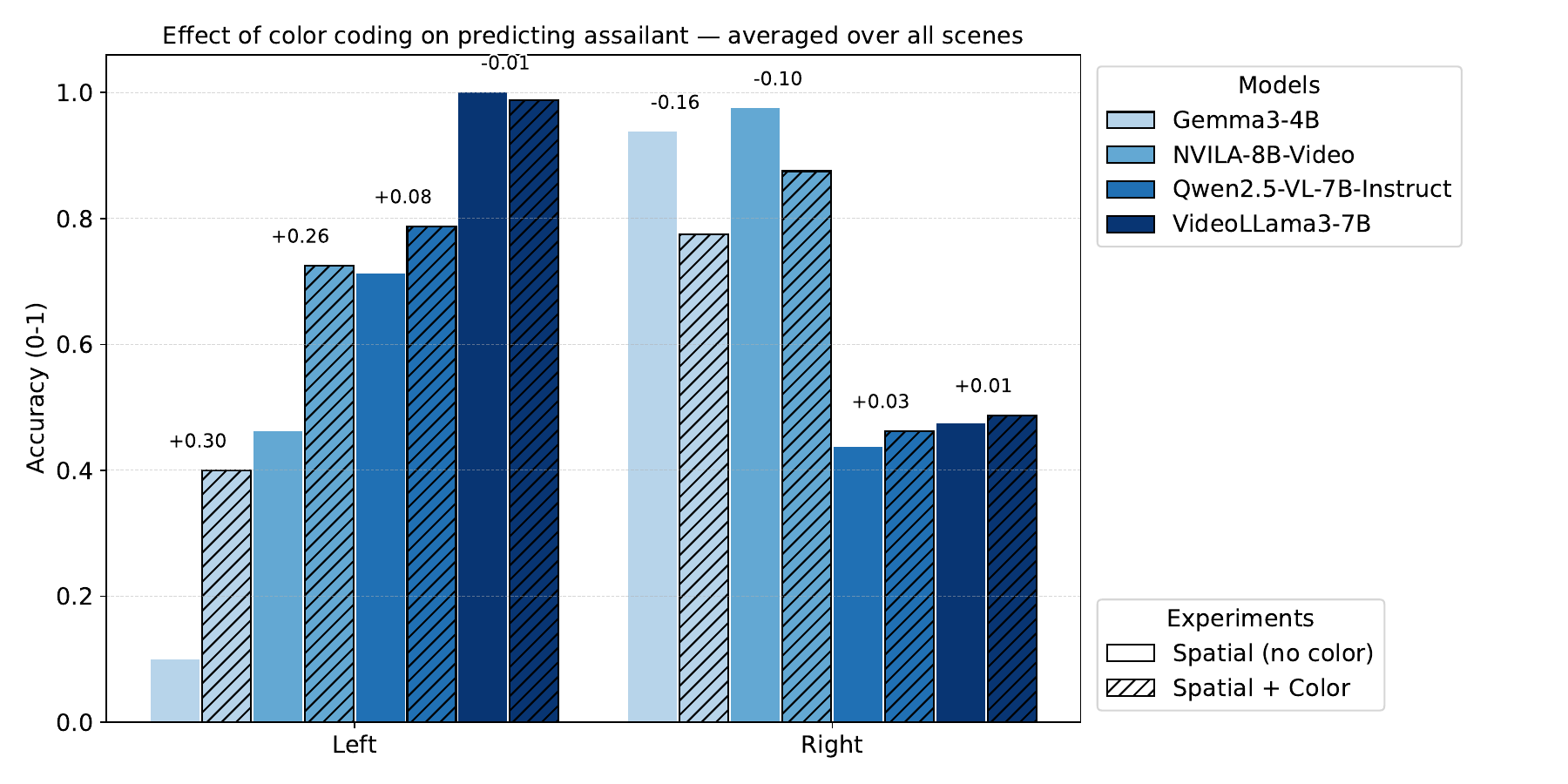}
    \caption{Role binding with and without stable color identifiers. Color substantially reduces left/right errors and role-flip sensitivity, isolating a linguistic anchoring failure.}
    \label{fig:left_right__spatial_vs_color}
\end{figure}

Figure~\ref{fig:left_right__spatial_vs_color} compares role binding with and without color identifiers. When only relative terms like “left” and “right” are available, models often stick to one interpretation of the viewpoint. Repeating the clips with fixed color identifiers (red/blue) reduces mistakes, indicating that many failures come from ambiguity in interpreting left/right, rather than from difficulties in tracking motion.

\begin{table}[t]
\centering
\small
\setlength{\tabcolsep}{6pt}
\begin{tabular}{lcccccc}
\toprule
\textbf{Model} & \textbf{Overall} & \textbf{Left} & \textbf{Middle} & \textbf{Right} & \textbf{HDRI} & \textbf{Zoom} \\
\midrule
Gemma3-4B                & 50.0 & 59.4 & 34.4 & 52.8 & 53.1 & 50.0 \\
NVILA-8B-Video           & 51.8 & 50.0 & 50.0 & 58.3 & 50.0 & 50.0 \\
Qwen2.5-VL-7B-Instruct   & 57.3 & 56.2 & 56.2 & 66.7 & 59.4 & 46.9 \\
VideoLLaMA3-7B           & 48.8 & 50.0 & 50.0 & 44.4 & 50.0 & 50.0 \\
\bottomrule
\end{tabular}
\caption{Following experiment (accuracy \%). Values are averages over two runs per model. Performance is near chance (50\%) overall, indicating that models largely fail to understand ``following'' as directional alignment rather than mere co-motion.}
\label{tab:following-summary}
\end{table}

Table~\ref{tab:following-summary} summarizes the following experiment, inspired by the concept of stalking. Accuracy hovers near chance across models and viewpoints, with modest improvements under zoom that are insufficient to close the gap. This task is both linguistically difficult (what is following in the exact same direction?) and visually difficult as the angle between trajectories is only 30$\degree$. 

\begin{figure}[h]
    \centering
    \includegraphics[width=1.0\linewidth]{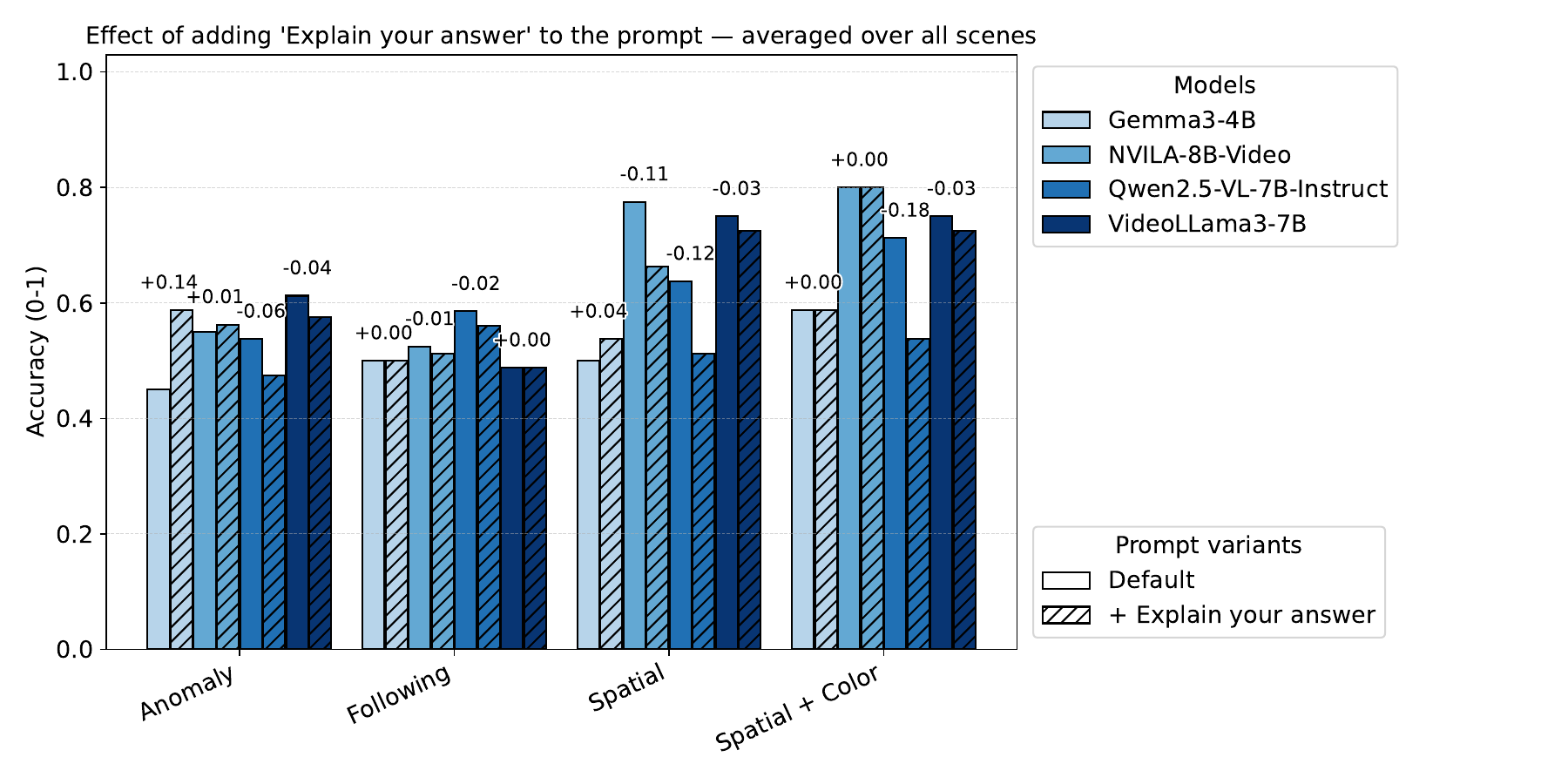}
    \caption{Impact of requiring explanations across experiments: beneficial for models with sparse outputs (e.g., Gemma) but detrimental for others (e.g., Qwen2.5-VL)}
    \label{fig:acc_across_prompts}
\end{figure}

Figure~\ref{fig:acc_across_prompts} shows that adding an explanation step in the prompt can help some models throw out less garbage, but it can also hurt other models. Gemma3 improved because it often generated too little text without a label, which leads to the text classifier guessing. 

Overall, these results suggest that current models capture only a rough sense of situations. They still have trouble with viewpoint-dependent role binding and with small differences in heading. A simple structural cue—using stable identifiers like clothing color—helps role binding without altering geometry or retraining. But heading alignment is tougher: models often treat similar movement as ‘following’ and overlook the importance of direction.

\section{Conclusion}
We presented a synthetic benchmark designed to stress-test spatial and situational reasoning in vision–language models through tightly controlled minimal pairs. Our results show that small VLMs consistently struggle—whether in separating violent from benign motion, resolving viewpoint-relative role binding, or interpreting the subtle trajectory cues that define following. The one partial exception is the spatial condition with explicit color identifiers, where models perform somewhat better, though still far from reliable. This suggests that lightweight structural priors can help, but they are insufficient on their own and point to the need for stronger forms of explicit grounding.

Beyond benchmarking, our study highlights a broader principle: explicit spatial grounding and reference stability are indispensable for reliable action recognition. Providing models with scene-centric coordinates and short-horizon trajectory cues offers a practical path toward robustness in safety-critical human–AI settings.

By releasing data and code, we aim not only to furnish diagnostic tools but also to seed exploration of structured priors that complement large-scale pretraining. Future directions include multi-party interactions, occlusion stress tests, and egocentric viewpoints (first-person perspectives from wearable or mobile cameras) — steps that will push models closer to the demands of real-world embodied intelligence.

\section{Discussion and Ethics}
This benchmark is a diagnostic instrument for research on spatial grounding and situational understanding. It is not a safety system and should not be used to make operational decisions. The dataset consists entirely of synthetic human animations rendered in virtual environments; no real identities are recorded, and no biometric information is collected. Although some clips depict fighting, we avoid graphic violence and keep durations short. To separate linguistic from perceptual confounds we sometimes assign stable colors to actors; these identifiers are introduced solely for analysis and are not intended to encode demographic categories. Synthetic scenes reduce privacy risk, but domain shift remains; models that perform well here may degrade in the wild. We release code, prompts, and data to enable reproduction under the licenses of the source assets (Unreal Engine/CitySample, Mixamo/Adobe, and Poly~Haven HDRIs), which prohibit training on the assets themselves. As the community develops stronger spatial priors, we recommend evaluating not only mean accuracy but also minimal-pair consistency, role-flip sensitivity, and robustness under zoom and viewpoint changes, which better capture grounded decisions that embodied systems must make.




\newpage
\bibliographystyle{plainnat}
\bibliography{bibliography}

\appendix

\newpage
\section{Prompts used in experiments} \label{app:prompts}

\begin{lstlisting}[breaklines=true, basicstyle=\ttfamily]
GUIDED_PROMPT_ANOMALY = """
You see two characters in the centre of the screen, one on the left and one on the right.
Your task is to decide if the clip shows a violent interaction (punching/striking/grappling = Anomaly) or non-violent motion (dancing/waving/standing = Normal).
Answer with "Anomaly" or "Normal"
"""

GUIDED_PROMPT_ANOMALY_SPATIAL = """
You see two characters in the centre of the screen, one on the Left and one on the Right.
Your task is to answer which person is attacking the other person.
Answer with "Left" or "Right"
"""

GUIDED_PROMPT_ANOMALY_SPATIAL_COLORED = """
You see two characters in the centre of the screen, one on the Left wearing Red clothing and one on the Right wearing Blue clothing.
Your task is to answer which person is attacking the other person.
Answer with "Left" or "Right"
"""

GUIDED_PROMPT_FOLLOWING = """
You see two characters in the centre of the screen, one on the Left and one on the Right.
Your task is to answer if one of the characters is following the other, the answer is True if the characters are walking in the exact same direction.
Answer with "True" or "False"
"""
\end{lstlisting}

The same prompts are used for the second set of experiments, where the only addition is: 
\begin{lstlisting}
    ", Explain your answer"
\end{lstlisting}

\section{Color added to tests} \label{app:color}

\begin{figure}[h]
    \centering
    \includegraphics[width=0.45\linewidth]{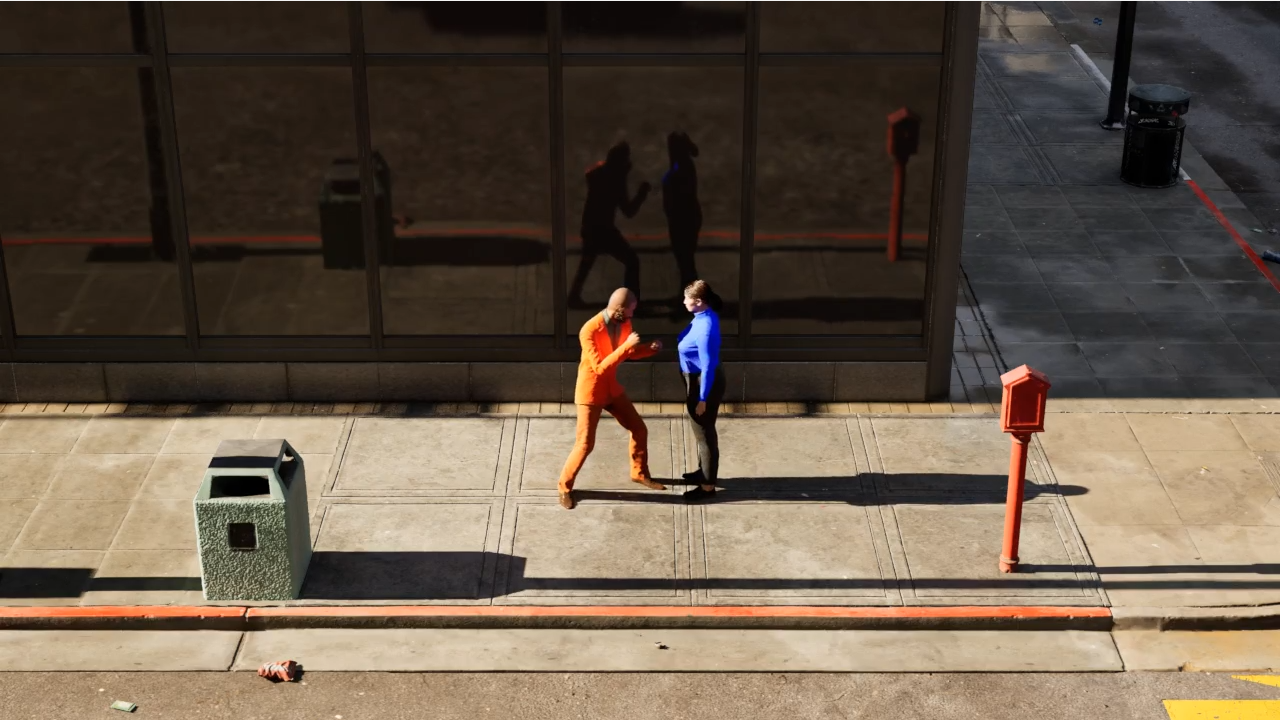}
    \includegraphics[width=0.45\linewidth]{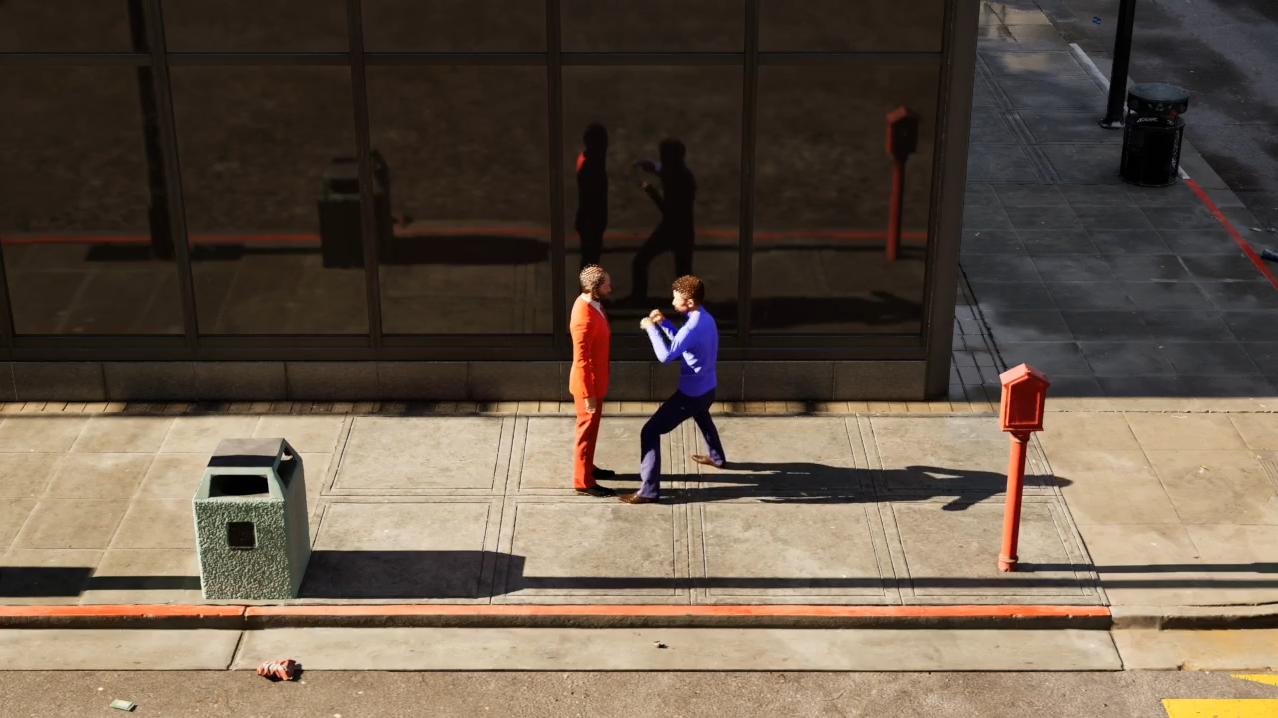}
    \caption{Adding color to spatial awareness tests, now the model can look at red vs blue instead of having to understand concepts like "left" and "right"}
    \label{fig:colored_spatial_example}
\end{figure}

\section{Unreal Engine setup} \label{app:unreal_setup}

\subsection{Positions of actors and cameras}
The camera in the Unreal Engine CitySample scene is positioned in the following way for the left, middle and right experiments:

\begin{table}[h!]
\centering
\caption{Camera transforms for left, middle, and right viewpoints.}
\label{tab:camera_transforms}
\begin{tabular}{lrrrrrr}
\toprule
\textbf{Camera} & \textbf{X (cm)} & \textbf{Y (cm)} & \textbf{Z (cm)} & \textbf{Roll (°)} & \textbf{Pitch (°)} & \textbf{Yaw (°)} \\
\midrule
Left   & -20600.0 & 34800.0 & 750.0 & 0.0 & -25.0 & 170.0 \\
Middle & -21200.0 & 34000.0 & 750.0 & 0.0 & -25.0 & 120.0 \\
Right  & -22100.0 & 34100.0 & 750.0 & 0.0 & -25.0 & 70.0  \\
\bottomrule
\end{tabular}
\end{table}

For the zoom experiment the focal length of the middle camera was increased from 28 to 50.

The characters where positioned as following:

\begin{table}[h!]
\centering
\caption{Character transforms (Unreal: Roll, Pitch, Yaw).}
\label{tab:character_transforms}
\begin{tabular}{lrrrrrr}
\toprule
\textbf{Character} & \textbf{X (cm)} & \textbf{Y (cm)} & \textbf{Z (cm)} & \textbf{Roll (°)} & \textbf{Pitch (°)} & \textbf{Yaw (°)} \\
\midrule
Character A & -21706.0 & 35104.0 & 171.0001 & 0.0 & 0.0 & 210.0 \\
Character B & -21920.0 & 35000.0 & 171.0001 & 0.0 & 0.0 & 30.0 \\
\bottomrule
\end{tabular}
\end{table}

\subsection{Animations used}
From \href{https://www.mixamo.com/}{Mixamo} we have used several animations:

\begin{table}[h!]
\centering
\begin{tabular}{l l l l}
\hline
\textbf{Dancing} & \textbf{Fighting} & \textbf{Idle} & \textbf{Walking} \\
\hline
Chicken\_Dance & Cross\_Punch & MTN\_N\_Idle & Walking (slowed to 4 seconds)\\
Silly\_Dancing & Hook\_Punch  & Old\_Man\_Idle \\
Wave\_Hip\_Hop\_Dance & Hook & Orc\_Idle \\
Ymca\_Dance    & Punching    & Standing\_W\_Briefcase\_Idle \\
\hline
\end{tabular}
\caption{Animations used from Mixamo, grouped by category.}
\label{tab:mixamo-animations}
\end{table}

\end{document}